\documentclass{article}

\usepackage{amsmath,amsfonts,bm}

\def\eqref#1{equation~\ref{#1}}

\def\1{\bm{1}}

\def\rvw{{\mathbf{w}}}

\def\rmE{{\mathbf{E}}}

\DeclareMathAlphabet{\mathsfit}{\encodingdefault}{\sfdefault}{m}{sl}
\SetMathAlphabet{\mathsfit}{bold}{\encodingdefault}{\sfdefault}{bx}{n}

\def\gA{{\mathcal{A}}}

\def\gC{{\mathcal{C}}}

\def\gG{{\mathcal{G}}}

\def\gJ{{\mathcal{J}}}

\def\gM{{\mathcal{M}}}
\def\gN{{\mathcal{N}}}

\def\gS{{\mathcal{S}}}

\DeclareMathOperator*{\argmax}{arg\,max}

\usepackage{subfiles}
\usepackage{microtype}
\usepackage{graphicx}
\usepackage{subfigure}
\usepackage{booktabs} 

\usepackage{hyperref}

\usepackage[accepted]{icml2023} 

\usepackage{amsmath}
\usepackage{amssymb}
\usepackage{mathtools}
\usepackage{amsthm}
\usepackage{stfloats}
\usepackage[capitalize,noabbrev]{cleveref}
\usepackage{enumitem}
\usepackage{graphicx} 
\usepackage{float}
\usepackage{array,booktabs,makecell,multirow}
\newcommand{\mb}[1]{\mathbf{#1}}
\newcommand{\mc}[1]{\mathcal{#1}}
\renewcommand{\algorithmiccomment}[1]{\STATE \textcolor{gray}{// #1}}
\renewcommand{\algorithmicrequire}{\textbf{Input:}}
\renewcommand{\algorithmicensure}{\textbf{Output:}}
\theoremstyle{plain}
\usepackage{thmtools, thm-restate}
\newtheorem{theorem}{Theorem}[section]

\newtheorem{lemma}[theorem]{Lemma}

\theoremstyle{definition}
\newtheorem{definition}[theorem]{Definition}

\theoremstyle{remark}

\usepackage[textsize=tiny]{todonotes}

\newcommand{\framework}{{COLE} framework\xspace}
\usepackage{xspace}
\newcommand{\algo}{$\text{COLE}_\text{SV}$\xspace} 
\usepackage{arydshln}
\DeclareUnicodeCharacter{2212}{-}

\icmltitlerunning{Cooperative Open-ended Learning Framework for Zero-shot Coordination}

\begin{document}

\twocolumn[
\icmltitle{Cooperative Open-ended Learning Framework for Zero-shot Coordination}



\icmlsetsymbol{equal}{*}

\begin{icmlauthorlist}
\icmlauthor{Yang Li}{equal,uom}
\icmlauthor{Shao Zhang}{equal,sjtu}
\icmlauthor{Jichen Sun}{sjtu}
\icmlauthor{Yali Du}{kcl}
\icmlauthor{Ying Wen}{sjtu}
\icmlauthor{Xinbing Wang}{sjtu}
\icmlauthor{Wei Pan}{uom}
\end{icmlauthorlist}

\icmlaffiliation{uom}{The University of Manchester}
\icmlaffiliation{sjtu}{Shanghai Jiao Tong University}
\icmlaffiliation{kcl}{King's College London}

\icmlcorrespondingauthor{Ying Wen}{ying.wen@sjtu.edu.cn}
\icmlcorrespondingauthor{Wei Pan}{wei.pan@manchester.ac.uk}

\icmlkeywords{Machine Learning, ICML}

\vskip 0.3in
]



\printAffiliationsAndNotice{\icmlEqualContribution} 


\begin{abstract}
Zero-shot coordination in cooperative artificial intelligence (AI) remains a significant challenge, which means effectively coordinating with a wide range of unseen partners. 
Previous algorithms have attempted to address this challenge by optimizing fixed objectives within a population to improve strategy or behaviour diversity. 
However, these approaches can result in a loss of learning and an inability to cooperate with certain strategies within the population, known as cooperative incompatibility.
To address this issue, we propose the \textbf{C}ooperative \textbf{O}pen-ended \textbf{LE}arning (\textbf{COLE}) framework, which constructs open-ended objectives in cooperative games with two players from the perspective of graph theory to assess and identify the cooperative ability of each strategy.  
We further specify the framework and propose a practical algorithm that leverages knowledge from game theory and graph theory. Furthermore, an analysis of the learning process of the algorithm shows that it can efficiently overcome cooperative incompatibility.
The experimental results in the Overcooked game environment demonstrate that our method outperforms current state-of-the-art methods when coordinating with different-level partners. Our demo and code are available at \url{https://sites.google.com/view/cole-2023/}. 
\end{abstract}

\section{Introduction}
Zero-shot coordination (ZSC) is a major challenge of cooperative AI to train agents that have the ability to coordinate with a wide range of unseen partners~\citep{Legg2007Universal,Hu2020OtherPlayFZ}.
The traditional method of self-play (SP)~\citep{SP} involves iterative improvement of strategies by playing against oneself. 
While SP can converge to an equilibrium of the game~\citep{Fudenberg1998Theory}, the strategies often form specific behaviours and conventions to achieve higher payoffs~\citep{Hu2020OtherPlayFZ}. 
As a result, a fully converged SP strategy may not be adaptable to coordinating with unseen strategies~\citep{lerer2018learning,Hu2020OtherPlayFZ}.

To overcome the limitations of SP, most ZSC methods focus on promoting strategic or behavioural diversity by introducing population-based training (PBT) to improve strategies' adaptive ability~\citep{HARL, Canaan2022Hanabi, MEP, TrajDi}. 
PBT aims to improve cooperative outcomes with other strategies in the population to promote zero-shot coordination with unseen strategies. 
This is achieved by maintaining a set of strategies to break the conventions of SP~\citep{SP} and optimizing the rewards for each pair in the population. 
Most state-of-the-art (SOTA) methods attempt to pre-train a diverse population~\citep{FCP, TrajDi} or introduce hand-crafted methods~\citep{Canaan2022Hanabi, MEP}, which are used to master cooperative games by optimizing fixed objectives within the population. 
These methods have shown to be efficacious in addressing intricate cooperative tasks such as Overcooked~\citep{HARL} and Hanabi~\citep{hanabi2020Nolann}.

However, when optimizing a fixed population-level objective, such as expected rewards within population~\citep{FCP,TrajDi,MEP} , the coordination ability of strategies within the population may not be improved. 
Specifically, while overall performance may improve, the coordination ability within the population may not be promoted in a simultaneous manner. 
This phenomenon, which we term ``\textit{cooperative incompatibility}", highlights the importance of considering the trade-offs between overall performance and coordination ability when attempting to optimize a fixed population-level objective.
 
In addressing the problem of cooperative incompatibility, we reformulate cooperative tasks as Graphic-Form Games (GFGs). 
In GFGs, strategies are characterized as nodes, with the weight of the edges between nodes representing the mean cooperative payoffs of the two associated strategies.
Additionally, by utilizing sub-graphs of GFGs referred to as preference Graphic-Form Games (P-GFGs), we are able to further profile each node's maximum cooperative payoff within the graph, enabling us to evaluate cooperative incompatibility and identify strategies that fail to collaborate.
Furthermore, we propose the Cooperative Open-ended LEarning (\textbf{COLE}) framework, which iteratively generates a new strategy that approximates the best response to the empirical gamescapes of P-GFGs. 
We have proved that the \framework can converge to the optimal strategy with a Q-sublinear rate when using in-degree centrality as the preference evaluation metric. 

To propose \framework to address the phenomenon of cooperative incompatibility, we implement a practical algorithm \algo by combining the \textbf{S}hapley \textbf{V}alue solution~\cite{shapley1971} with our GFG. 
\algo comprises a simulator, a solver, and a trainer, designed to master cooperative tasks with two players specifically.
The solver, utilizing the development of the intuitive solution concept Shapley value, evaluates the adaptive ability of strategies and calculates the cooperative incompatibility distribution.
The trainer aims to approximate the best responses to the cooperative incompatibility distribution mixture in the most recent population. 
To evaluate the performance of the~\algo, we conducted experiments in Overcooked, a two-player cooperative task environment~\citep{HARL}.
 We evaluated the adaptive ability of \algo by testing its performance against different level partners - middle-level and expert unseen partners. 
 The middle-level partner is a commonly used behavior cloning model~\citep{HARL}, and the expert partners are strategies of current methods, i.e., SP, PBT, FCP, and MEP, and our proposed \algo.
 The results of the experiments showed that \algo outperforms the recent SOTA methods in both evaluation protocols. 
 Additionally, through the analysis of GFGs and P-GFGs, the learning process of \algo revealed that the framework efficiently overcomes cooperative incompatibility. 
The contributions in this paper can be summarized as follows.
\begin{itemize}
    \item We introduce the concept of Graphic-Form Games (GFGs) and Preference Graphic-Form Games (P-GFGs) to intuitively reformulate cooperative tasks, which allows for a more efficient evaluation and identification of cooperative incompatibility during learning.
    \item We develop the concept of graphic-form gamescapes to help understand the objective and present the \framework to iteratively approximate the best responses preferred by most others.
    \item We prove that the algorithm will converge to the optimal strategy, and the convergence rate will be Q-sublinear when using in-degree preference centrality. Empirical experiments in the game Overcooked verify the proposed algorithm's effectiveness compared to SOTA methods.
\end{itemize}

\section{Related Works}

\textbf{Zero-shot coordination.} The goal of zero-shot coordination (ZSC) is to train a strategy that can coordinate effectively with unseen partners~\citep{Hu2020OtherPlayFZ}. 
Self-play~\citep{SP,HARL} is a traditional method of training a cooperative strategy, which involves iterative improvement of strategies by playing against oneself, but develops conventions between players and does not cooperate with other unseen strategies~\citep{lerer2018learning,Hu2020OtherPlayFZ}. 
Other-play~\citep{Hu2020OtherPlayFZ} is proposed to break such conventions by adding permutations to one of the strategies.
However, this approach may be reduced to self-play if the game or environment does not have symmetries or has unknown symmetries. 
Another approach is population-based training (PBT)~\citep{PBT,HARL}, which trains strategies by interacting with each other in a population.
However, PBT does not explicitly maintain diversity and thus fails to coordinate with unseen partners\citep{FCP}. 

Recent research has focused on training robust strategies that use diverse populations of strategies~\citep{FCP,TrajDi,MEP} to achieve the goal of ZSC.
Fictitious co-play (FCP)~\citep{FCP} obtains a population of periodically saved checkpoints during self-play training with different seeds and then trains the best response to the pre-trained population. 
TrajeDi~\citep{TrajDi} also maintains a pre-trained self-play population but encourages distinct behavior among the strategies. 
The maximum entropy population (MEP)~\citep{MEP} method proposes population entropy rewards to enhance diversity during pre-training. It employs prioritized sampling to select challenging-to-collaborate partners to improve generalization to previously unseen policies. 
Furthermore, methods such as MAZE~\citep{Xue2022Heter} and CG-MAS~\citep{mahajan2022generalization} have been proposed to improve generalization ability through coevolution and combinatorial generalization.
In this paper, we propose a \framework that could dynamically identify strategies that fail to coordinate due to cooperative incompatibility and continually poses and optimizes objectives to overcome this challenge and improve adaptive capabilities.

\textbf{Open-ended learning.}
Another related area of research is open-ended learning, which aims to continually discover and approach objectives~\citep{Srivastava2012Comtinually, Team2021OpenEndedLL,meier2022open}.
In MARL, most open-ended learning methods focus on zero-sum games, primarily posing adaptive objectives to expand the frontiers of strategies~\citep{NIPS2017_3323fe11,psrorn,mcaleer2020pipeline,yaodong_diverse,liu2021towards,mcaleer2022self}.
In the specific context of ZSC, the MAZE method~\citep{Xue2022Heter} utilizes open-ended learning by maintaining two populations of strategies and partners and training them collaboratively throughout multiple generations. 
In each generation, MAZE pairs strategies and partners from the two populations and updates them together by optimizing a weighted sum of rewards and diversity. 
This method co-evolves the two populations of strategies and partners based on naive evaluations such as best or worst performance with strategies in partners.
Our proposed method, \framework, combines GFGs and P-GFGs in open-ended learning to evaluate and identify the cooperative ability of strategies to solve cooperative incompatibility efficiently with theoretical guarantee.

\begin{figure*}[ht!]
\includegraphics[width=0.95\linewidth]{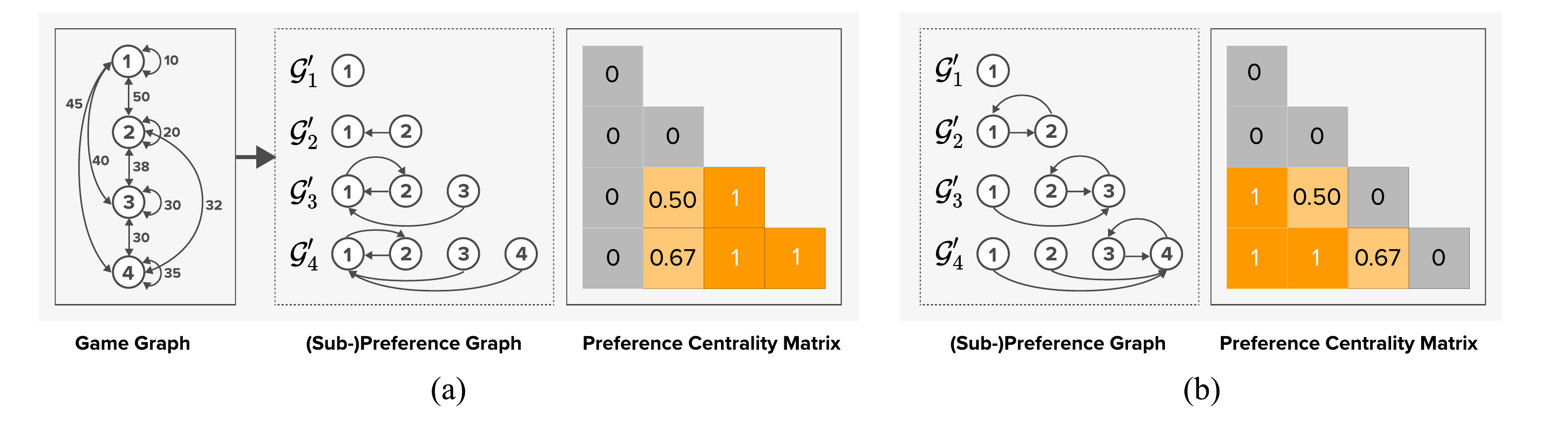}
\centering
\caption{ {The Game Graph, (sub-) preference graph and corresponding preference centrality matrix.}
The (sub-) preference graphs are for all four iterations in the training process, and the corresponding preference in-degree centrality matrix is based on them.
 As can be observed in the $\gG^\prime_3$ and $\gG^\prime_4$, the newly updated strategies fail to be preferred by others and have centrality values of 1, despite an increase in the mean of rewards with all others. 
 In \textit{(b)}, we illustrate an ideal learning process in which a newly generated strategy can achieve higher outcomes with all previous strategies.
}
\label{fig:game_graph}
\end{figure*}


\section{Preliminaries}
\paragraph{Normal-form Game.}
A two-player normal-form game is defined as a tuple $(N, \gA, \rvw)$, where $N=\{1,2\}$ is a set of two players, indexed by $i$, $\gA=\gA_1 \times \gA_2$ is the joint action space, and $\rvw=(w_1, w_2)$ with $w_i: \gA \rightarrow \mathbb{R}$ is a reward function for the player $i$. 
In a two-player common payoff game, two-player rewards are the same, meaning $w_1(a_1,a_2)=w_2(a_1,a_2)$ for $a_1,a_2 \in \gA$.

\paragraph{Empirical Game-theoretic Analysis (EGTA), Empirical Game and Empirical Gamescape.} EGTA is the study of finding meta-strategies based on experience with prior strategies~\citep{Walsh2002Analyzing, Karl2018Generalised}. 
An empirical game is built by discovering strategies and meta-reasoning about exploring the strategy space~\citep{NIPS2017_3323fe11}.
Furthermore, empirical gamescapes (EGS) are introduced to represent strategies in functional form games geometrically~\citep{psrorn}.
Given a population $\gN$ of $n$ strategies, the empirical gamescapes is often defined as 
$
    \gG := \{\text{convex mixture of rows of}~\gM \},
$
where $\gM$ is the empirical payoff table recording the expected outcomes for each joint strategy.

\paragraph{Cooperative Theoretic Concepts.} We consider a set of players $\gN = \{1, \dots, n\}$, where a coalition is denoted as a subset of players $\gN$, symbolized by $C \subseteq \gN$. The player set $\gN$ is also known as the grand coalition. A characteristic function game, denoted by $G$, consists of a pair $(N, v)$, where $N = \{1, \dots, n\}$ represents a finite, non-empty set of agents, and $v: 2^N \rightarrow \mathbb{R}$ is the characteristic function. This function assigns a real number $v(C)$ to each coalition $C \subseteq N$, with the number $v(C)$ typically considered as the coalition's value.

Shapley Value~\cite{shapley1971} is one of the important solution concepts for characteristic function games \cite{chalkiadakis2011computational,Bezalel2007intro}.
The Shapley Value aims to distribute fairly the collective value, like the rewards and cost of the team across individuals by each player's contribution.
Taking into account a coalition game $(\gN,v)$ with a strategy set $\gN$ and characteristic function $v$, the Shapley Value of a player $i\in \gN$ could be obtained by 

\begin{equation}
SV(i)=\frac{1}{n!}\sum_{\pi\in\Pi_\mathcal{N}} v(P_i^\pi \cup \{i\}) - v(P_i^\pi),
\label{eq:SV}
\end{equation}
where $\pi$ is one of the one-to-one permutation mappings from $\gN$ to itself in the permutation set $\Pi$ and $\pi(i)$ is the position of player $i \in \gN$ in permutation $\pi$. $P_i^\pi=\{j\in \gN | \pi(j)<\pi(i)\}$ is the set of all predecessors of $i$ in $\pi$.



\paragraph{Centrality in Graph Theory.}  Node Centrality is a graph theory concept that quantifies a node's relative importance or influence within a network. It is a measure of how central a node is to the overall network structure, with higher centrality values indicating greater importance. 
Degree Centrality is one of the simplest centrality concepts, based on the number of edges connected to a node~\cite{FREEMAN1978215}. 
In directed graphs, it can be further divided into in-degree (number of incoming edges) and out-degree (number of outgoing edges) Centrality.

PageRank~\cite{page1999pagerank} is a centrality measure used to rank web pages in search engine results by estimating the relative importance of each page in a hyperlink network. PageRank(WPG)~\cite{Xing2004Weighted} is an extension of the original PageRank algorithm that considers edge weights in addition to the basic structure of the network. This modification allows the algorithm to handle networks where the strength of connections between nodes varies, such as in citation networks or social networks where the influence of nodes may differ.
The formula of WPG is given as follows:
\begin{equation}
    \sigma(u)=(1-d) + d \sum_{v\in B(u)} \sigma(v) \frac{I_u}{\sum_{p\in R(v)} I_p} \frac{O_u}{\sum_{p\in R(v)}O_p}, 
    \label{eq:wpg}
\end{equation}   
where $d$ is the damping factor set to $0.85$, $B(u)$ is the set of nodes that point to $u$, $R(v)$ denotes the nodes to which $v$ is linked, and $I, O$ are the degrees of inward and outward of the node, respectively.

\section{Cooperative Open-Ended Learning}
In this section, we first introduce graphic-form games to intuitively reformulate cooperative games, then create an open-ended learning framework to solve cooperative incompatibility and further improve zero-shot adaptive ability.

\subsection{Graphic-Form Games (GFGs)} 
It is important to evaluate cooperative incompatibility and identify those failed-to-collaborate strategies to conquer cooperative incompatibility.
Therefore, we propose graphic-form games (GFGs) to reformulate normal-form cooperative games from the perspective of game theory and graph theory, which is the natural development of empirical games~\citep{psrorn}.
The definition of GFG is given below.

\begin{definition}[Graphic-Form Game]
    Given a set of parameterized strategies $\gN=\{1,2,\cdots,n\}$, a two-player graphic-form game (GFG) is a tuple $\gG = (\gN, \rmE, \rvw)$, which could be represented as a directed weighted graph.
    $\gN,\rmE,\rvw$ are the set of nodes, edges, and weights, respectively.
    Given an edge $(i,j)$, $\rvw(i,j)$ represents the expected results of $i$ playing with $j$.
     The graphic representation of GFG is called a game graph.
\end{definition}
The payoff matrix of $\gG$ is denoted as $\gM$, where $\gM(i,j)=\rvw(i,j), \forall i,j \in \gN$.
Our goal is to improve the upper bound of other strategies' outcomes in the cooperation within the population, which implies that the strategy should be preferred over other strategies.

Moreover, we propose preference graphic-form games (P-GFGs) as an efficient tool to analyze the current learning state, which can profile the degree of preference for each node in GFGs.
Specifically, P-GFG is a subgraph of GFG, where each node only retains the out-edge with maximum weight among all out-edges except for its self-loop.
\begin{definition}[Preference Graphic-Form Game]
Given a graphic-form game $(\gN, \rmE, \rvw)$, the Preference Graphic-Form Game (P-GFG) could be defined as $\gG^\prime = \{\gN,\rmE^\prime, \rvw\}$, where $\rmE^\prime=\{(i,j) | \argmax_j \rvw(i, j), \forall j\in \{\gN\backslash i\}, \forall i \in \gN\}$ is the set of edges. The graphic representation of P-GFG is called a preference graph.
\end{definition}

To deeply investigate the learning process, we further introduce the \textit{sub-preference graphs} based on P-GFGs, which aim to reformulate previous learning states and analyze the learning behavior of the algorithm.
Suppose that there is a set of sequentially generated strategies $\gN_n=\{1,2,\cdots,n\}$, where the index also represents the number of iterations for simplicity.
For each previous iteration $i<n$, the sub-preference game form graph is denoted as $\{\gN_i, \rmE^\prime_i,\mb{w}_i\}$, where $\gN_i=\{1,2,\cdots,i\}$ is the set of strategies in iteration $i$, and $\rmE^\prime_i, and\ \mb{w}_i$ are the corresponding edges and weights.

The semantics of the preference graph is that a strategy or node $i$ prefers to play with the tailed node to achieve the highest results.
In other words, the more in-edges one node has, the more cooperative ability this node can achieve.
Ideally, if one strategy can adapt well to all others, all the other strategies in the preference graph will point to this strategy.
To evaluate the adaptive ability of each node, the centrality concept is introduced into the preference graph to evaluate how a node is preferred.
\begin{definition}[Preference Centrality]
    Given a P-GFG $\{\gN,E^\prime, \rvw\}$, preference centrality of $i\in \gN$ is defined as,
    $$
    \eta(i)=1- \operatorname{norm}(d_i),
    $$
    where $d_i$ is a graph centrality metric to evaluate how the node is preferred, and $\operatorname{norm}(\cdot):\mathbb{R}\rightarrow [0,1]$ is a normalization function.
\end{definition}

Note that the $d$ is a kind of centrality that could evaluate how much a node is preferred.
A typical example of $d$ is the centrality of degrees, which calculates how many edges point to the node. 

Fig.~\ref{fig:game_graph} is an example of a common payoff game, showing the game graph, (sub-)preference graphs, and the preference centrality matrix for four sequentially generated strategies.
Note that in the corresponding sub-preference graphs, the updated strategies fail to improve the outcome of others after the second iteration,
and the preference centrality matrix also shows the same results.
The example shows an existing cooperative incompatibility that presents as the value of $\eta$ is kept at 1 in the matrix, meaning no nodes want to collaborate with the updated strategies.
Ideally, all the other strategies should prefer latest strategy (Fig.~\ref{fig:game_graph} (b)) which means the 
monotonic improvement of cooperative ability.
\begin{figure}[t!]
\centering
\includegraphics[width=0.8\linewidth]{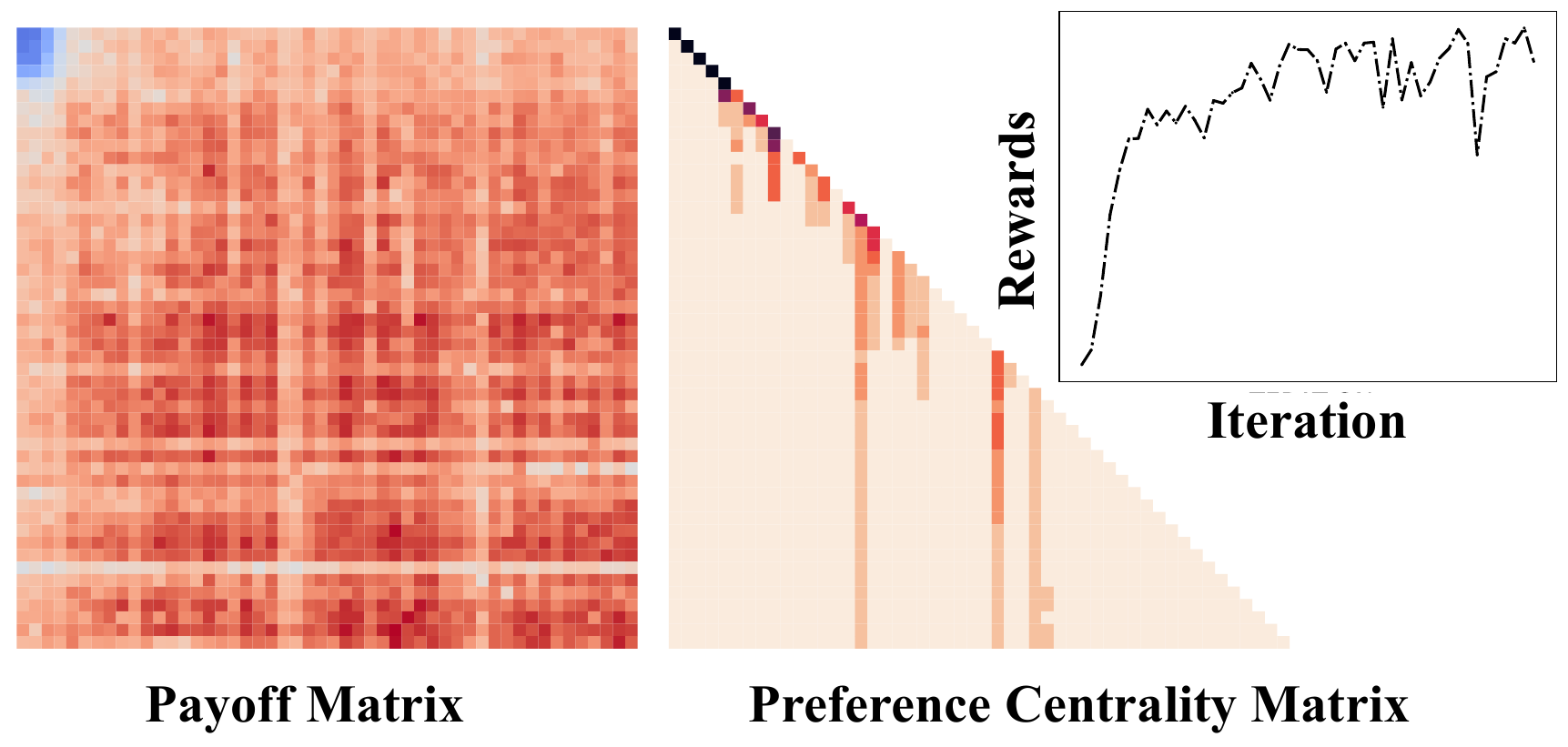}
\caption{
The payoff matrix of each strategy during training and the corresponding preference centrality matrix of the MEP algorithm in the Overcooked. The darker the color in the payoff matrix, the higher the rewards. The darker the color in the preference centrality matrix, the lower the centrality value, and the more other strategies prefer it.
}
\label{fig:mep_eta}
\end{figure}


Moreover, the analysis of the MEP algorithm, as shown in Fig.~\ref{fig:mep_eta}, discloses a cooperative incompatibility in the learning process in Overcooked environment~\citep{HARL}.
In the preference indegree centrality matrix, a strategy is preferred by more strategies if its color is darker.
In the learning process of MEP, although the mean rewards are always improving (as shown in the upper-right of Fig.~\ref{fig:mep_eta}), serious cooperative incompatibility problems occur after a period of training, where more strategies prefer to play with some previous strategies with a darker color rather than new strategies to obtain higher rewards.
\begin{figure*}[ht!]
    \centering
\includegraphics[width=0.85\linewidth]{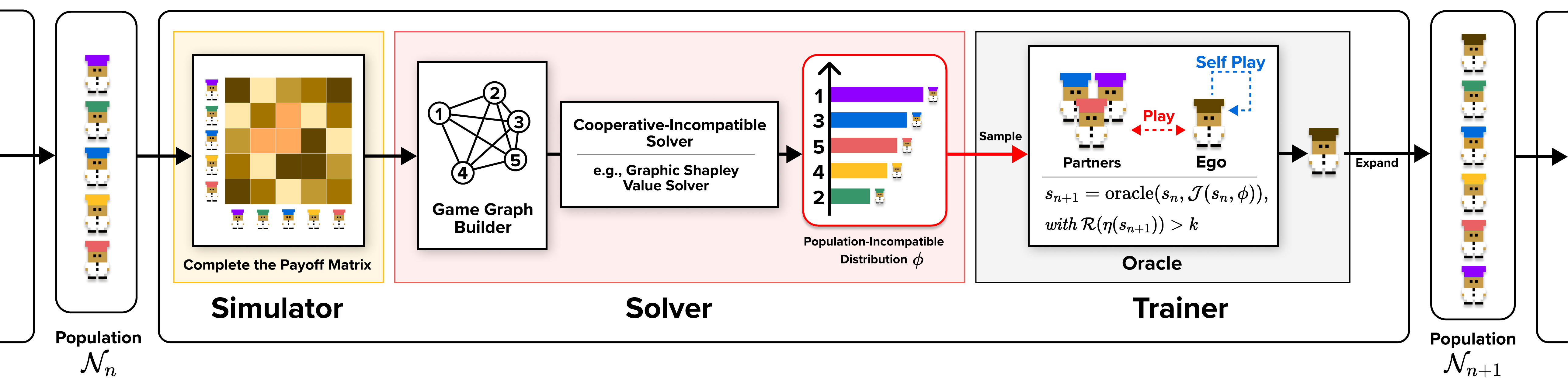}
    \caption{
    An overview of one generation in \framework: The solver derives the cooperative incompatible distribution $\phi$ using a cooperative incompatibility solver, which can be any algorithm that evaluates cooperative contribution. The trainer then approximates the relaxed best response by optimizing individual and cooperative compatible objectives. The oracle's training data is generated using partners selected based on the cooperative incompatibility distribution and the agent's strategy. Finally, the approximated strategy $s_{n+1}$ is added to the population, and the next generation begins.
    }
    \label{fig:cole}
\end{figure*}
\subsection{Cooperative Open-Ended Learning Framework}
To tackle cooperative incompatibility by understanding the objective, we develop empirical gamescapes ~\citep{psrorn} for GFGs, which geometrically represent strategies in graphic-form games.
This provides us with a geometric representation of player strategies within a GFG $\{\gN, \rmE, \rvw\}$, thereby capturing the diversity and adaptive capacity of cooperative strategic behaviour.

However, learning directly with EGS to cooperate with these well-collaborated strategies is inefficient in improving adaptive ability.
To conquer cooperative incompatibility, the natural idea is to learn with the mixture of cooperative incompatible distribution on the most recent population $\gN$.
Given a population $\gN$, we present \textit{cooperative incompatible solver} to assess how strategies collaborate, especially with those strategies that are difficult to collaborate with.
The solver derives the cooperative incompatible distribution $\phi$, where strategies that do not coordinate with others have higher probabilities.

We also optimize the cooperative incompatible mixture over the individual objective, which is the cumulative self-play rewards to improve the adaptive ability with expert partners.
To simplify, we name it the individual and cooperative incompatible mixture (IPI mixture).
We use an approximate oracle to approach the best response over the IPI mixture.
Given strategy $s_n$, the oracle returns a new strategy $s_{n+1}$ :
$
    s_{n+1} = \operatorname{oracle}(s_{n+1}, \gJ(s_{n}, \phi)),
$
with $\eta(s_{n+1})=0$ , if possible. 
$\gJ$ is the objective function as follows,
\begin{equation}
    \label{eq:obj}
    \gJ(s_n,\phi) = \mathbb{E}_{p\sim \phi}\rvw(s_n,p) + \alpha \rvw(s_n,s_n),
\end{equation}
where $\alpha$ is the balance hyperparameter.
The objective consists of the cooperative compatible objective and the individual objective.
The cooperative compatible objective aims to train the best response to those failed-to-collaborate strategies, and the individual objective aims to improve the adaptive ability with expert partners.
We call the best response the local best-preferred strategy in the population if $\eta(s_{n+1})=0$. 

However, arriving at the local best-preferred strategy with $\eta(s_{n+1})=0$ is hard or even impossible.
Therefore, we seek to approximate the local best-preferred strategies by relaxing the local best strategy to the strategy whose preference centrality ranks top $k$.
The approximate oracle could be rewritten as 
\begin{equation}
    s_{n+1} = \operatorname{oracle}(s_n, \gJ(s_n, \phi)), with\ \mathcal{R}(\eta(s_{n+1}))>k,
    \label{eq:oracle_approx}
\end{equation}
where $\mathcal{R}$ is the ranking function.

We extend the approximated oracle to open-ended learning and propose \framework (Fig.~\ref{fig:cole}).
The \framework iteratively updates new strategies that approximate the local best-preferred strategies to the cooperative incompatible mixture and the individual objective.
The simulator completes the payoff matrix with the newly generated strategy and others in the population.
The solver aims to derive the cooperative incompatible distribution of the Game Graph builder and the cooperative-incompatible solver.
The trainer uses the oracle to approximate the local best-preferred strategy to the cooperative incompatible mixture and individual objective and outputs a newly generated strategy added to the population for the next generation.

Although we relax the local best-preferred strategy to the strategy in the top $k$ centrality in the constraint, \framework still converges to a local best-preferred strategy with zero preference centrality.
Formally, the local best-preferred strategy convergence theorem is given as follows.
 \begin{restatable}{theorem}{FirstTHM}
    \label{thm: converge}
Let $s_0\in \gS$ be the initial strategy and $s_i=\operatorname{oracle}(s_{i-1})$ for $i \in \mathbb{N}$.
Under the effective functioning of the approximated oracle as characterized by Eq.~\ref{eq:oracle_approx}, we can say that the sequence $\{s_i\}$ for ${i\in \mathbb{N}}$ could converge to a local optimal strategy $s^*$, i.e., the local best-preferred strategy.
 \end{restatable}
 \begin{proof} 
 See Appendix~\ref{appendix:proofs_thm}.
\end{proof}
Besides, if we choose in-degree centrality as the preference centrality function, the convergence rate of \framework is Q-sublinear.
\begin{restatable}{corollary}{FirstLEMMA}
    \label{lemma: converge_rate}
Let $\eta: \gG^\prime \rightarrow \mathbb{R}^n$ be a function that maps a P-GFG to its in-degree centrality, the convergence rate of the sequence $\{s_i\}$ is Q-sublinear concerning $\eta$.
 \end{restatable}
\begin{proof}
See Appendix~\ref{appendix:proofs_corollary}.
\end{proof}

\section{Practical Algorithm}
\label{sec:COOL}
To address common-payoff games with two players, we implemented \algo, where SV refers to \emph{Shapley Value}, based on \framework that can overcome cooperative incompatibility and improve zero-shot coordination capabilities, focusing on the solver and trainer components.
As shown in Fig.~\ref{fig:cole}, at each generation, \algo inputs a population $\gN$ and generates an approximate local best-preferred strategy added to $\gN$ to expand the population.
The simulator calculates the payoff matrix $\gM$ for the input population $\gN$. 
Each element $\gM(i,j)$ for $i,j\in \gN$ represents the cumulative rewards of the players $i$ and $j$ at both starting positions. 
The solver evaluates and identifies failed-to-collaborate strategies by calculating the incompatible cooperative distribution. 
To effectively evaluate the cooperative ability of each strategy with all others, we incorporate weighted PageRank (WPG)~\citep{Xing2004Weighted} from graph theory into the Shapley Value to evaluate adaptability, particularly with failed-to-collaborate strategies. 
The trainer then approximates the local best-preferred strategy over the recent population.


\subsection{Solver: Graphic Shapley Value}
To approximate the local best-preferred strategies over the recent population and overcome cooperative incompatibility, we need to calculate the cooperative incompatible distribution as the mixture.
In this paper, we combine the Shapley Value~\citep{shapley1971} solution, an efficient single solution concept for cooperative games to assign the obtained team value across individuals, with our GFG to evaluate and identify the strategies that did not cooperate.
To apply the Shapley Value, we define an additional characteristic function to evaluate the value of the coalition.
Formally, given a coalition $C\subseteq \gN$, we have the following:
\begin{equation}
  v(C) = \mathbb{E}_{i\sim C,j\sim C}\sigma(i)\sigma(j)\rvw(i,j),
\end{equation}
where $\sigma$ is a mapping function that evaluates how badly a node performs on its game graph.
We use the characteristic function to evaluate the coalition value of how it could cooperate with those hard-to-collaborate strategies.

We take the inverse of WPG~\citep{Xing2004Weighted} on the game graph as the metric $\sigma$.
WPG is proposed to assess the popularity of a node in a complex network, as given in Eq.~\ref{eq:wpg}.
Therefore, the metric $\sigma$ evaluates how unpopular a node is and is equal to the inverse of the WPG value.

Then we calculate the Shapley Value of each node by taking a characteristic function in equation~\ref{eq:SV}, named the graphic Shapley Value.
We utilize the Monte Carlo permutation sampling~\citep{Castro2009PolynomialCO} to approximate the Shapley Value, which can reduce the computation complexity from exponential time to linear time.
After inverting the probabilities of the graphic Shapley Value, we get the cooperative incompatible distribution $\phi$, where strategies that fail to collaborate with others have higher probabilities.
We provide the Graphic Shapley Value algorithm in Appendix~\ref{appedix:solver}.

\subsection{Trainer: Approximating local best-preferred Strategy}

The trainer takes the cooperative incompatible distribution $\phi$ as input and samples its teammates to learn to approach the local best-preferred strategy on the IPI mixture.

Recall the oracle for $s_n$ : 
$
    s_{n+1} = \operatorname{oracle}(s_{n+1}, \gJ(s_{n}, \phi)),
$
with $\mathcal{R}(\eta(s_{n+1}))>k$.
\algo aims to optimize the local best-preferred strategy over the IPI mixture. 
The $\gJ(s_{n}, \phi)$ is the joint objective that consists of individual and cooperative compatible objectives.
The individual objective aims to improve the performance within itself and promote the adaptive ability with expert partners, formulated as follows:
$
    \gJ_i(s_n) = \rvw(s_n,s_n),
$
where $s_n$ is the strategy named ego strategy that needs to optimize in generation $n$.

And the cooperative compatible objective aims to improve cooperative outcomes with those failed-to-collaborate strategies:
$
    \gJ_{c} = \mathbb{E}_{p\sim \phi}\rvw(s_n,p),
$
where the objective is the expected rewards of $s_n$ with cooperative incompatible distribution-supported partners.
$\rvw$ estimates and records the mean cumulative rewards of multiple trajectories and starting positions.
The expectation can be approximated as:
\begin{equation}
\gJ_{c} = \sum^b_{p\sim \phi}\phi(p)\rvw(s_t,p),
\end{equation}
where $b$ is the number of sampling times.

\label{sec:solver}
\begin{algorithm}[t!]
\caption{\algo Algorithm}
\label{alg:cool}
\begin{algorithmic}[1]
\STATE \algorithmicrequire population $\gN_{0}$, the sample times $a,b$ of $\gJ_i,\gJ_c$, hyperparameters $\alpha,k$

\FOR{$t = 1,2,\cdots, $}
    \COMMENT{Step 1: Completing the payoff matrix}
    \STATE $\gM_n \leftarrow \operatorname{Simulator}(\gN_t)$
    \COMMENT{Step 2: Solving the cooperative incompatibility distribution}
    \STATE $\phi = \operatorname{Graphic\ Shapley\ Value}(\gN_t)$ by Algorithm~\ref{algo:solver}
    \COMMENT{Step 3: Approximate the local best-preferred strategy}
    \STATE $\gJ = \sum^b_{p\sim \phi}\phi(p)\rvw(s_t,p) + \alpha \sum^a\rvw(s_t,s_t)$, where $s_t=\gN_t(t)$, $\phi$ is updated each time by Eq~\ref{eq:SUCG}
    \STATE {$s_{t+1} = \operatorname{oracle}(s_t, \gJ)$} with $\mathcal{R}(\eta(s_{n+1}))>k$
    \COMMENT{Step 4: Expand the population}
    \STATE $\gN_{t+1} = \gN_{t} \cup \{s_{t+1}\}$ 
    \ENDFOR
\end{algorithmic}
\end{algorithm}
\setlength{\textfloatsep}{3mm}

To balance exploitation and exploration as the learning continues, we present the Sampled Upper Confidence Bound for Game Graph (SUCG) that combines the Upper Confidence Bound (UCB) and GFG to control the sampling for more strategies with higher probabilities or new strategies.
Additionally, we view the SUCG value as the probability of sampling teammates instead of using the maximum item in typical UCB algorithms.
Specifically, in the game graph, we keep the information on the times that a node has been visited.
Therefore, the probability of each node considers both the Shapley Value and visiting times, denoted as $\hat{p}$.
The SUCG for any node $u$ in $\gN$ could be calculated as follows:
\begin{equation}
    \hat\phi(u) = \phi(u) + c\frac{\sqrt{\sum_{i\in \gN} \mb{N}(i)}}{1+\mb{N}(u)},
    \label{eq:SUCG}
\end{equation}
where $c$ is a hyperparameter that controls the degree of exploration and $\mb{N}(i)$ is the visit times of node $i$.
SUCG could efficiently prevent \algo from generating data with a few fixed strategies that did not cooperate, which could lead to a loss of adaptive ability.

We conclude the \algo as Algorithm~\ref{alg:cool}.
Moreover, to verify the influence of different ratios of two objectives, we denote \algo with different ratios as 0:4, 1:3, 2:2, and 3:1.
Specifically, \algo with $a:b$ represents different partner sampling ratios for the combining objective, where $a$ is the corresponding times to generate data using self-play for the individual objective, and $b$ is the number of sampling times in $\gJ_c$.
For example, \algo 1:3 represents that the ego agent is trained by using self-play once and with partners sampled from the cooperative incompatible distribution three times to generate data and update objectives.
\section{Experiments}
\label{sec:exp}
\subsection{Environment and Experimental Setting}

\begin{figure}[t!]
    \centering
    \includegraphics[width=0.95\linewidth]{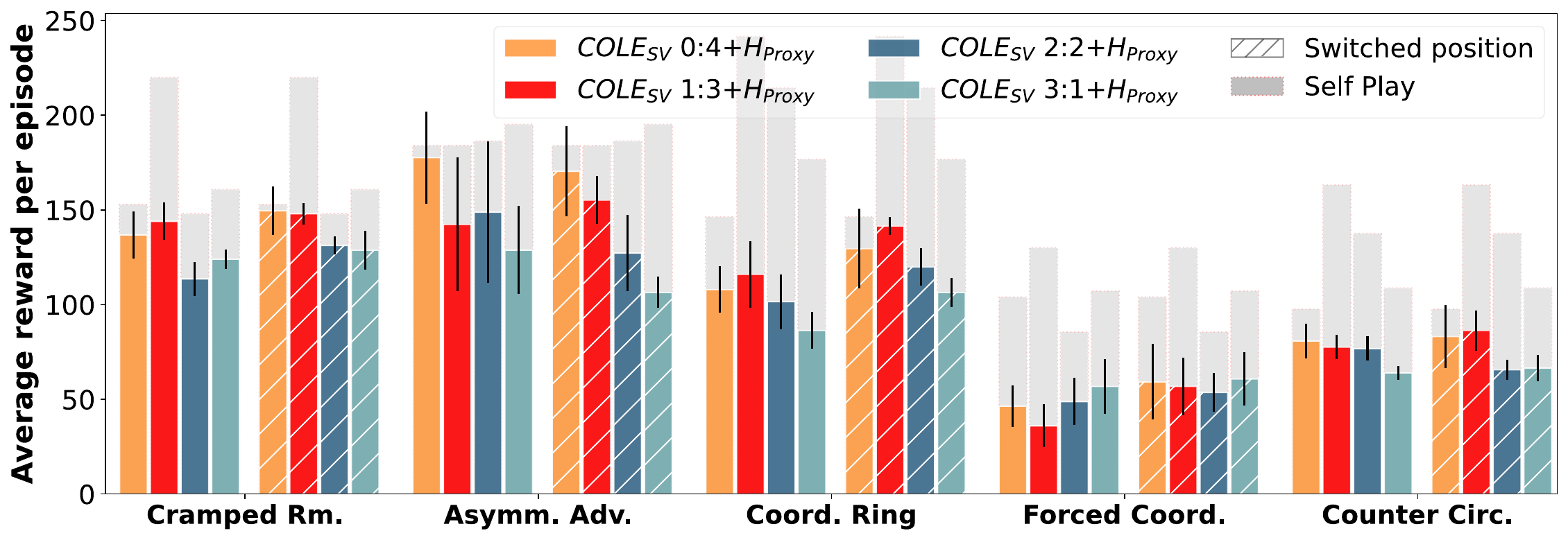}
    \caption{{The result of the combining objectives' effectiveness evaluation.}
    Mean episode rewards over 400 timesteps trajectories for \algo s with different objective ratios 0:4, 1:3, 2:2, and 3:1, paired with the unseen middle-level partner $H_{proxy}$.
    The gray bars behind present the rewards of self-play.
    }
    \label{fig:exp_ablation}
\end{figure}

In this paper, we conduct a series of experiments in the Overcooked environment~\citep{HARL,charakorn2020investigating,knott2021evaluating}.
The details of the Overcooked environment can be found in Appendix~\ref{appedix:layouts}.
We construct evaluations with different ratios between individual and cooperative compatible objectives, such as 0:4, 1:3, 2:2, and 3:1. 
These studies demonstrate the effectiveness of optimizing both individual and cooperative incompatible goals. 
We also compare our method with other methods, including self-play~\citep{SP,HARL}, PBT~\citep{PBT,HARL}, FCP~\citep{FCP}, and MEP~\citep{MEP}, all of which use PPO~\citep{schulman2017proximal} as the RL algorithm. 
To thoroughly assess the ZSC ability, we evaluated the algorithms with unseen middle-level and expert partners. 
We use the human proxy model $H_{proxy}$ proposed by Carroll et al.\citep{HARL} as middle-level partners and the models trained with baselines and \algo as expert partners. 
The rewards' mean is recorded as the performance of each method in collaborating with expert teammates. 
In the case study, we analyze the learning process of \algo, which shows that our method overcomes cooperative incompatibility. 
Furthermore, we visualize the trajectories with different ratios and play with expert teammates to analyze how the ratios affect the learned strategies. 
Appendix~\ref{appendix:cole} and Appendix~\ref{appendix:base} give details of the implementation of \algo and baselines.

\subsection{Combining Objectives' Effectiveness Evaluation}

This section evaluated the effectiveness of different objective ratios, including 0:4, 1:3, 2:2, and 3:1 of two objectives.
We divided each training batch into four parts, the ratio indicating the proportion of data generated by self-play and data generated by playing with strategies from the cooperative incompatible distribution. 
We omitted the 4:0 ratio as it would result in the framework degenerating into self-play.
Fig.~\ref{fig:exp_ablation} shows the mean rewards of episodes over 400 timesteps of gameplay when paired with the unseen middle-level partner $H_{proxy}$ \citep{HARL}. 
We found that \algo with ratios 0:4 and 1:3 achieved better performance than the other ratios. 
In particular, \algo, with a ratio of 1:3, outperformed the other methods in the Cramped Room, Coordination Ring, and Counter Circuit layouts. 
On the Forced Coordination layout, which is particularly challenging for cooperation due to the separated regions, all four ratios performed similarly on average across different starting positions.  
Interestingly, when paired with the middle-level partner, \algo with only the cooperative compatible objective (ratio 0:4) performed better on the Asymmetric Advantages and Forced Coordination layouts.
We discuss this phenomenon further in Section \ref{different_levels}.
The effectiveness evaluations indicate that combining individual and cooperatively compatible objectives is crucial to improving performance with unseen partners.
In general, we choose the ratio of 1:3 as the best choice.

\begin{figure}[t!]
    \centering
    \includegraphics[width=0.95\linewidth]{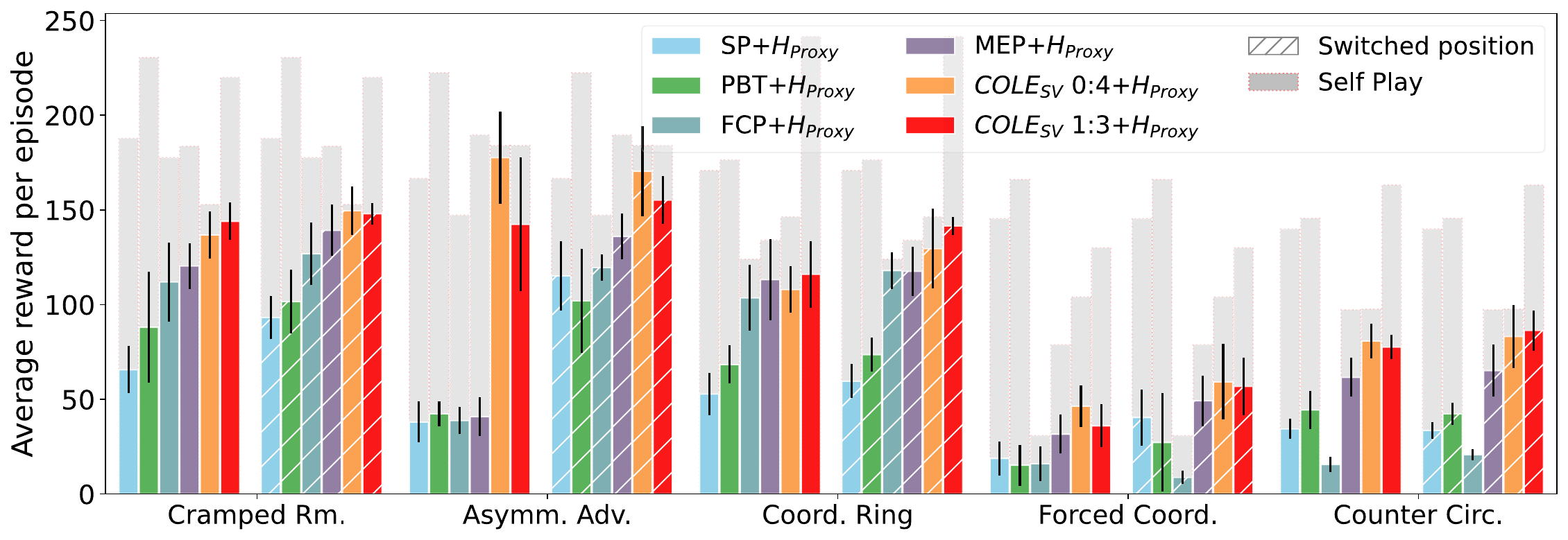}
    \caption{{Performance with middle-level partners.}
    The performance of \algo with middle-level partners is presented in terms of mean episode rewards over 400 timesteps trajectories for different objective ratios of 0:4 and 1:3, when paired with the unseen middle-level partner $H_{proxy}$. The results include the mean and standard error over five different random seeds. The gray and hashed bars indicate the rewards obtained when playing with themselves and the performance when starting positions are switched.
    }
    \label{fig:main_exp}
\end{figure}
\subsection{Evaluation with Different Levels of Partners}
\label{different_levels}

To thoroughly evaluate the zero-shot cooperative ability of all methods, we adopted two sets of evaluation protocols. 
The first protocol involves playing with a trained human model $H_{proxy}$ trained in behavior cloning.
However, due to the quality and quantity of human data used for behavior cloning to train the human model is limited, the capabilities of the human proxy model can only be classified as middle-level. 
Therefore, we use an additional evaluation protocol to coordinate with unseen expert partners. 
We selected the best models of our reproduced baselines and \algo 0:4 and 1:3 as expert partners.

Fig.~\ref{fig:main_exp} presents the performance of SP, PBT, MEP, and \algo with 0:4 and 1:3 when cooperating with middle-level partners. 
We observed that different starting positions on the left and right in asymmetric layouts resulted in significant performance differences for the baselines. 
For example, in the Asymmetric Advantages, the cumulative rewards of all baselines in the left position were nearly one-third of those in the right position. 
On the contrary, \algo performed well at the left and right positions.

As shown in Fig.~\ref{fig:main_exp}, \algo outperforms other methods in all five layouts when paired with the middle-level partner-human proxy model. 
Interestingly, \algo 0:4 with only the cooperatively compatible objective achieves better performance than \algo 1:3 on some layouts, such as Asymmetric Advantages. 
However, the self-play rewards of \algo 0:4 are much lower than \algo 1:3 and even other baselines. 
Furthermore, the performance with unseen experts of \algo 0:4 as shown in Table~\ref{tab:exp_expert}, is sometimes lower than the baselines.
We visualize the trajectories in the evaluation at the expert level and provide further analysis to explain this situation in Appendix~\ref{Trajectory}.

\begin{table}[t!]
\caption{Performance with expert partners. Mean episode rewards over 1 min trajectories for baselines and \algo with ratio 0:4, 1:3.
  Each column represents a different expert group, in which the result is the mean reward for each model playing with all others.
  }
\label{tab:exp_expert}
\begin{center}
\resizebox{\linewidth}{!}{%
\begin{sc}
    \begin{tabular}{lcccccc}
\toprule
\multirow{2}{*}{\textbf{Layout}} &\multirow{2}{*}{\textbf{Ratio}} & \multicolumn{4}{c}{\textbf{Baselines}}  &\multirow{2}{*}{\textbf{COLEs}}\\
\cline{3-6}\
 && \textbf{SP} & \textbf{PBT} & \textbf{FCP} & \textbf{MEP} &  \\
\midrule
\multirow{2}{*}{\textbf{Cramped Rm.}} &0:4&
153.00 & 198.50  & {199.83 } & 178.83 & 169.76 \\&1:3& 
165.67 & 209.83 & 207.17 & 196.83 & \textbf{212.80}\\
\hline
\multirow{2}{*}{\textbf{Asymm.Adv.}} &0:4&
108.17  & 164.83 & 175.50 & 179.83& \textbf{182.80}
\\&1:3&
108.17 & 161.50 & 172.17 & {179.83} & 178.80\\
 \hline
\multirow{2}{*}{\textbf{Coord. Ring}}&0:4&
132.00 & 106.83 & {142.67} & 130.67  & 118.08
\\&1:3&
133.33 & 158.83 & 144.00  & 124.67 &\textbf{166.32}\\
 \hline
\multirow{2}{*}{\textbf{Forced Coord.}} &0:4&
~~58.33 & ~~61.33 & ~~50.50  & ~~{79.33} &  ~~46.40\\&1:3&
~~61.50  & ~~70.33 & ~~62.33  & ~~38.00  &~~\textbf{86.40}\\
 \hline
\multirow{2}{*}{\textbf{Counter Circ.}}&0:4&
~~44.17  & ~~48.33 & ~~60.33& ~~21.33 & ~~{90.72}\\
&1:3&
~~65.67  & ~~64.00  & ~~46.50  & ~~76.67  &  \textbf{105.84}
\\
\bottomrule
\end{tabular}
\end{sc}
}
\end{center}
\end{table}
Table~\ref{tab:exp_expert} presents the outcomes of each method when cooperating with expert partners. 
Each column in the table represents different expert groups, including four baselines and one \algo with a ratio of 0:4 or 1:3. 
The last column, labeled ``COLEs," represents the mean rewards of the corresponding \algo when working with other baselines. 
The table displays the mean cumulative rewards of each method when working with all other models in the expert group. 
The results indicate that \algo 1:3 outperforms the baselines and \algo 0:4, except in the layout of Asymmetric Advantages.
In the Asymmetric Advantages, \algo 0:4 only achieved a four-point victory over \algo 1:3, which can be considered insignificant considering the margin of error. 
In the other four layouts, the rewards obtained by \algo 1:3 while working with expert partners are significantly higher than those of \algo 4:0 and the baselines.

Our results suggest that \algo 1:3 has a stronger adaptive ability with different levels of partners. Furthermore, individual objectives are crucial in zero-shot coordination with expert partners. 
In conclusion, \algo 1:3 is more robust and flexible in real-world scenarios when working with partners of different levels.

\subsection{Effectively Conquer Cooperative Incompatibility}
\label{casestudy} 
\begin{figure}[t!]
\centering    \includegraphics[width=0.9\linewidth]{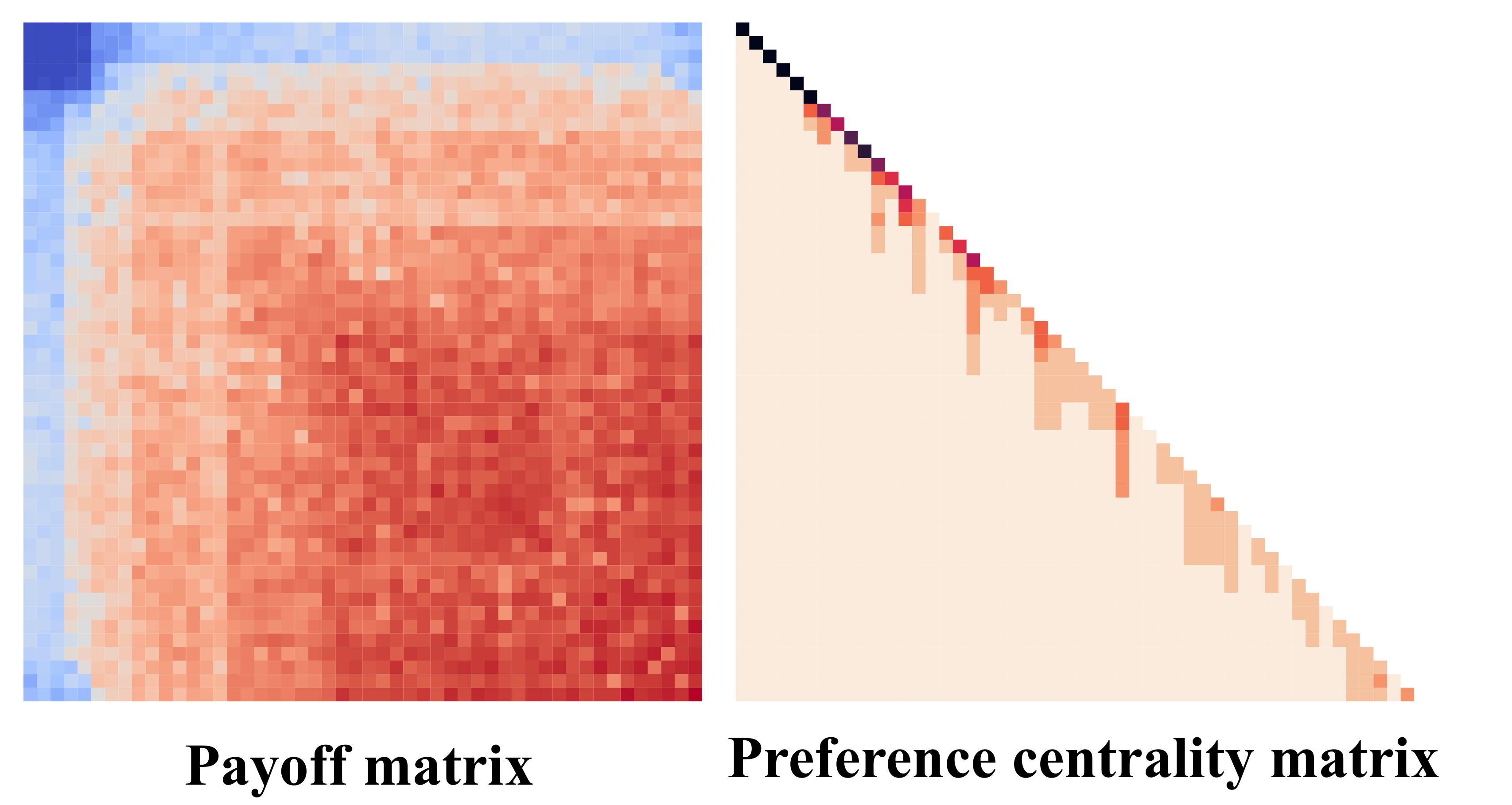}
    \caption{
    {The learning process analysis of \algo 1:3.}
The darker-colored element on the left represents higher rewards, while the darker-colored element on the right represents lower centrality. The clustering of darker-colored areas around the diagonal on the right indicates that the new strategy adopted in each generation is preferred by most strategies, thus overcoming the cooperative incompatibility.
}
\label{fig:cole_analysis}
\end{figure}
In our analysis of the learning process of \algo 1:3 in the Overcooked environment, as shown in Fig.~\ref{fig:cole_analysis}, we observe that the method effectively overcomes the problem of cooperative incompatibility. 
The figure on the left in Fig.~\ref{fig:cole_analysis} shows the payoff matrix of 50 uniformly sampled checkpoints during training, with the upper left corner representing the starting point of training. 
Darker red elements in the payoff matrix indicate higher rewards. 
The figure on the right displays the centrality matrix of preferences, which is calculated by analyzing the learning process. 
Unlike the payoff matrix, the darker elements in the centrality matrix indicate lower values, indicating that more strategies prefer them in the population. 
As shown in the figure, the darker areas cluster around the diagonal of the preference centrality matrix, indicating that most of the others prefer the updated strategy of each generation. 
Thus, we can conclude that our proposed \algo effectively overcomes the problem of cooperative incompatibility.

\section{Conclusion}
In this paper, we propose graphic-form games and preference graphic-form games to intuitively reformulate cooperative games, which can efficiently evaluate and identify cooperative incompatibility. Furthermore, we develop empirical gamescapes for GFG to understand the objectives and present \framework to iteratively approximate the best response preferred by most others over the most recent population. 
Theoretically, we prove that \framework converges to the optimal strategy preferred by all others. 
Furthermore, if we choose the in-degree centrality as the preference centrality function, the convergence rate would be Q-sublinear. 
Empirically, our experiments on the Overcooked environment show that our algorithm \algo outperformed SOTA ones and that \algo efficiently overcame cooperative incompatibility.

\textbf{Limitations and Future Work.} Our practical algorithm, \algo, incorporates the Shapley Value as a tool and develops the Graphic Shapley Value to scrutinize cooperative incompatibility. Despite our use of Monte Carlo permutation sampling to mitigate computational complexity, this remains a challenge. As such, future efforts should focus on enhancing the efficiency of the Graphic Shapley Value solver and investigating alternative solvers for evaluating cooperative incompatibility. Besides, our framework is restricted to two-player games. Therefore, future research should also aim to extend the COLE framework and develop corresponding algorithms for more complex multi-player games.

\section*{Acknowledgements}

Yang Li is supported by the China Scholarship Council (CSC) Scholarship. 
The Shanghai Jiao Tong University team is supported by the National Key R\&D Program of China (2022ZD0114804), National Natural Science Foundation of China (No.62106141), and Shanghai Sailing Program (21YF1421900).
The authors also thank Xihuai Wang for his kind assistance and advice.

\bibliography{reference}
\bibliographystyle{icml2023}

\clearpage
\onecolumn
\appendix
\section{Reproducibility Statement}
Our demo is provided in \url{https://sites.google.com/view/cole-2023/}.

Besides, more reproducibility information could be found in the appendix.
\begin{itemize}
\vspace{-3mm}
\setlength{\itemsep}{1pt}
\setlength{\parskip}{1pt}
\setlength{\parsep}{1pt}
    \item The detailed pseudocode of the Graphic Shapley Value Solver is provided in Appendix~\ref{appedix:solver}.
    \item In Appendix~\ref{appedix:layouts}, we introduce the details of the experimental environment - the Overcooked game.
    \item The implementation and hyperparameters used in experiments are in Appendix~\ref{appendix:cole}.
\end{itemize}

\section{Proofs of Theorem~\ref{thm: converge}}
\label{appendix:proofs_thm}
\FirstTHM*
\begin{proof}

According to the definition of the local best-preferred strategy, the local optimal strategy is the node with zero preference centrality ($\eta$). Therefore, we need to prove that the value of $\eta$ will approach zero.

{
Let $\eta_t$ denote the centrality value of the preference of the updated strategy $s_t$ in generation $t$, where $0\leq \eta \leq 1$. 
We first remark on the RL oracle defined in Eq.~\ref{eq:oracle_approx}:
$
    s_{n+1} = \operatorname{oracle}(s_n, \gJ(s_n, \phi)),
\ with\ 
\mathcal{R}(\eta(s_{n+1}))>k.
$
Under the assumption of the approximated oracle functioning effectively, it follows that the preference centrality $\eta_t$ associated with generated strategy $s_t$ at generation $t$ resides among the first $k$ positions when arranged in ascending order of centrality values.
For simplicity, We define a group $g_t$ at generation $t$ as the set of strategies with the lowest $k$ preference centrality values. 
\begin{lemma}
\label{lemma:group}
    Provided that the approximated oracle is functioning effectively, the maximal preference centrality value, denoted as $\eta_{g_t}$, within group $g_t$ is expected to exhibit a diminishing trend with each successive generation. 
    \end{lemma}
\begin{proof}
    In light of the approximated oracle's definition, the preference centrality value of the strategy $s_t$, generated at generation $t$, will occupy one of the first $k$ positions when sorted in ascending order based on preference centrality values. Consequently, the initial $k$ strategies of group $g_t$ will undergo an update, in which the strategy ranking $k$-th with the highest preference centrality value is substituted by either the $(k-1)$-th strategy or the newly generation strategy, $s_t$. Under any given conditions, the maximum preferential value within the group will experience a reduction.
\end{proof}
}

{
Let $\eta_{g_t}$ denote the largest preference centrality in the group $g$.
Thus, we can derive the subsequent equation based on Lemma~\ref{lemma:group}. 
\begin{equation} 
\eta_{g_t} = \eta_{g_{t-1}} - \epsilon_{t-1}, 
\end{equation} 
where $\epsilon_{t-1}$ is a positive value and $0< \epsilon \leq \eta_{g_{t-1}}$.
By further simplifying the equation, we have
\begin{equation} 
\begin{aligned} 
\eta_{g_{t}} &= \eta_{g_{t-1}} - \epsilon_{t-1},\\ &=\eta_{g_{t-1}} -\alpha_{t-1} \eta_{g_{t-1}},\\ &=\beta_{t-1} \eta_{g_{t-1}}, \end{aligned} 
\end{equation} 
where the second line employs $\eta_{g_{t-1}}$ to substitute the residual term, with the adjustment parameters $0 < \alpha_{t-1} \leq 1$ and $\beta_{t-1} = 1 - \alpha_{t-1}.$}

{
Assuming that the centrality value of the preference in the initial step is $0\leq \eta_0 \leq 1$, we can recursively calculate the following formula: 
\begin{equation} \begin{aligned} \eta_{g_t} &=\beta_{t-1} \eta_{g_{t-1}},\\ &=\beta_{t-1} \beta_{t-2} \eta_{g_{t-2}},\\ &=\cdots, \\ &=\prod_{i=0}^{t-1} \beta_i \times \eta_{g_0}. 
\end{aligned} 
\label{eq:iter_eta} \end{equation} For any $\beta \in {\beta_0, \cdots, \beta_{t-1}}$, we have $1>\beta\geq 0$. In addition, we set $\beta_t$ as a very small positive number if $\eta_t=0$.
Furthermore, we ascertain that the coefficient $\beta$ is not consistently zero. This is due to the fact that $\beta = 0$ would imply a preference centrality of zero for the strategy, which is not universally attainable within the context of the RL oracle. This very limitation underpins our rationale for introducing the approximated RL oracle.
Thus, we can conclude that $\eta_t$ will approach zero within the population as outlined in~\eqref{eq:iter_eta}.} 

{
Through this proof, we have substantiated that under the effective functioning of the RL oracle as characterized by Eq.~\ref{eq:oracle_approx}, the sequence ${s_i}$ is progressing towards the zero of preference centrality within the population. That is, the sequence is converging to a strategy denoted by $s^*$, which represents a locally best-preferred solution.
}
\end{proof}

\section{Proof of Corollary~\ref{lemma: converge_rate}}
\label{appendix:proofs_corollary}
\FirstLEMMA*
\begin{proof}
    In Theorem~\ref{thm: converge}, we have proved that the strategies generated by the \framework~will converge to the local best-preferred strategy.
When we use the in-degree centrality function as $\eta$, the preference centrality function can be rewritten as:
\begin{equation}
        \eta(i) = 1-\frac{I_i}{n-1},
    \end{equation}
where $I_i$ is the in-degree of node $i$ and $n$ is the size of the strategy set $\gN$.

{
    Therefore, we have
    \begin{equation}
\begin{aligned}
\label{eq:proof_1}
&\lim\limits_{t \to \infty} \frac{|\eta_{t+1} - 0|}{|\eta_{t} - 0|} \\
        = &\lim\limits_{t \to \infty} \frac{\eta_{t+1}}{\eta_{t}} \\
        =& \lim\limits_{t \to \infty} \frac{1-\frac{I_{t+1}}{t}}{1-\frac{I_{t}}{t-1}} \\
        = &\lim\limits_{t \to \infty} \frac{t-1}{t} \frac{t - I_{t+1}}{t-I_t-1} \\
        =&\lim\limits_{t \to \infty} \frac{t - I_{t+1}}{t-I_t-1} \\
        =&1
\end{aligned}
\end{equation}
}

Therefore, using the in-degree centrality, we can conclude that the \framework~will converge to the local optimal strategy at a Q-sublinear rate.
\end{proof}

\section{Graphic Shapley Value Solver Algorithm }
\label{appedix:solver}
Algorithm~\ref{algo:solver} gives the detailed steps of the graphic Shapley value solver in Section~\ref{sec:solver}.

\begin{algorithm}[ht]
\caption{Graphic Shapley Value Solver Algorithm}
\label{algo:solver}
\begin{algorithmic}[1]
\STATE \algorithmicrequire: population $\gN$, the number of Monte Carlo permutation sampling $k$, the size of negative population

\STATE Initialize $\phi = \mb{0}_{|\gC|}$
\FOR{$(1,2,\cdots, k)$}
    \STATE $\pi \longleftarrow \textit{Uniformly sample from } \Pi_\gC$, where $\Pi_\gC$ is permutation set
    \FOR{$i\in \mc{N}$}
    \COMMENT{Obtain predecessors of player $i$ in sampled permutation $\pi$}
    \STATE $S_\pi(i) \longleftarrow \{j\in \mc{N} | \pi(j)<\pi(i)\}$
    \COMMENT{Update incompatibility weights}
    \STATE $\phi_i\longleftarrow {\phi}_i + \frac{1}{k}({v(S_{\pi}(i)\cup \{i\})-v(S_{\pi}(i)))}$
    \ENDFOR
    \ENDFOR
\STATE $\phi \longleftarrow \phi/\sum\phi$
\STATE $\phi \longleftarrow (1-\phi)/\sum(1-\phi)$
\STATE \algorithmicensure: $\phi$
\end{algorithmic}
\end{algorithm}


\section{Overcooked Environment}
\label{appedix:layouts}
In this paper, we conduct a series of experiments in the Overcooked environment~\citep{HARL,charakorn2020investigating,knott2021evaluating}, which is proposed for the coordination challenge, to verify the performance of \algo. 
As a two-player common payoff game, each player controls one chef in a kitchen to cook and serve soup, which results in a reward of 20 for the team. We test our codes on five different layouts: Cramped Room, Asymmetric Advantages, Coordination Ring, Forced Coordination, and Counter Circuit. 

\begin{figure}[ht] \centering    
\subfigure[{\scriptsize Cramped Room}] {\includegraphics[height=65pt]{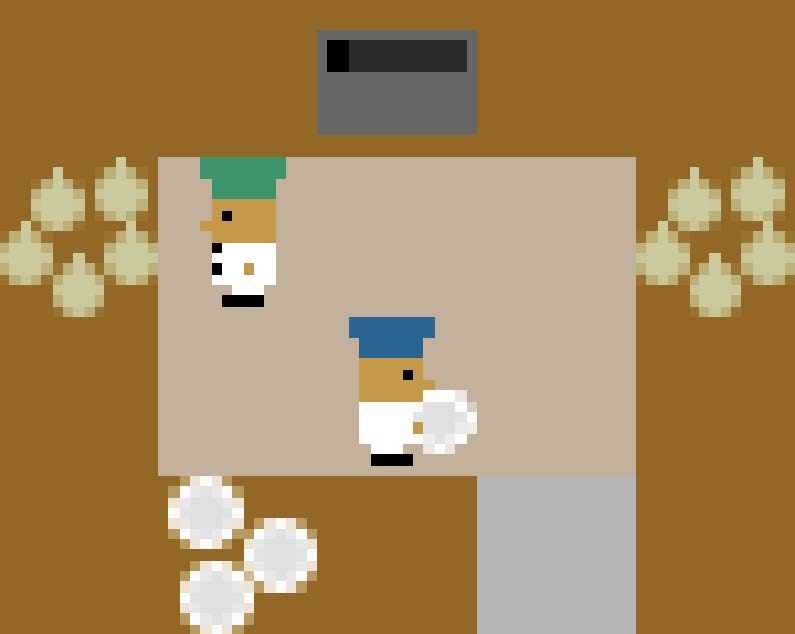}
}   
\subfigure[{\scriptsize Asymmetric Advantages}] {   
    \includegraphics[height=65pt]{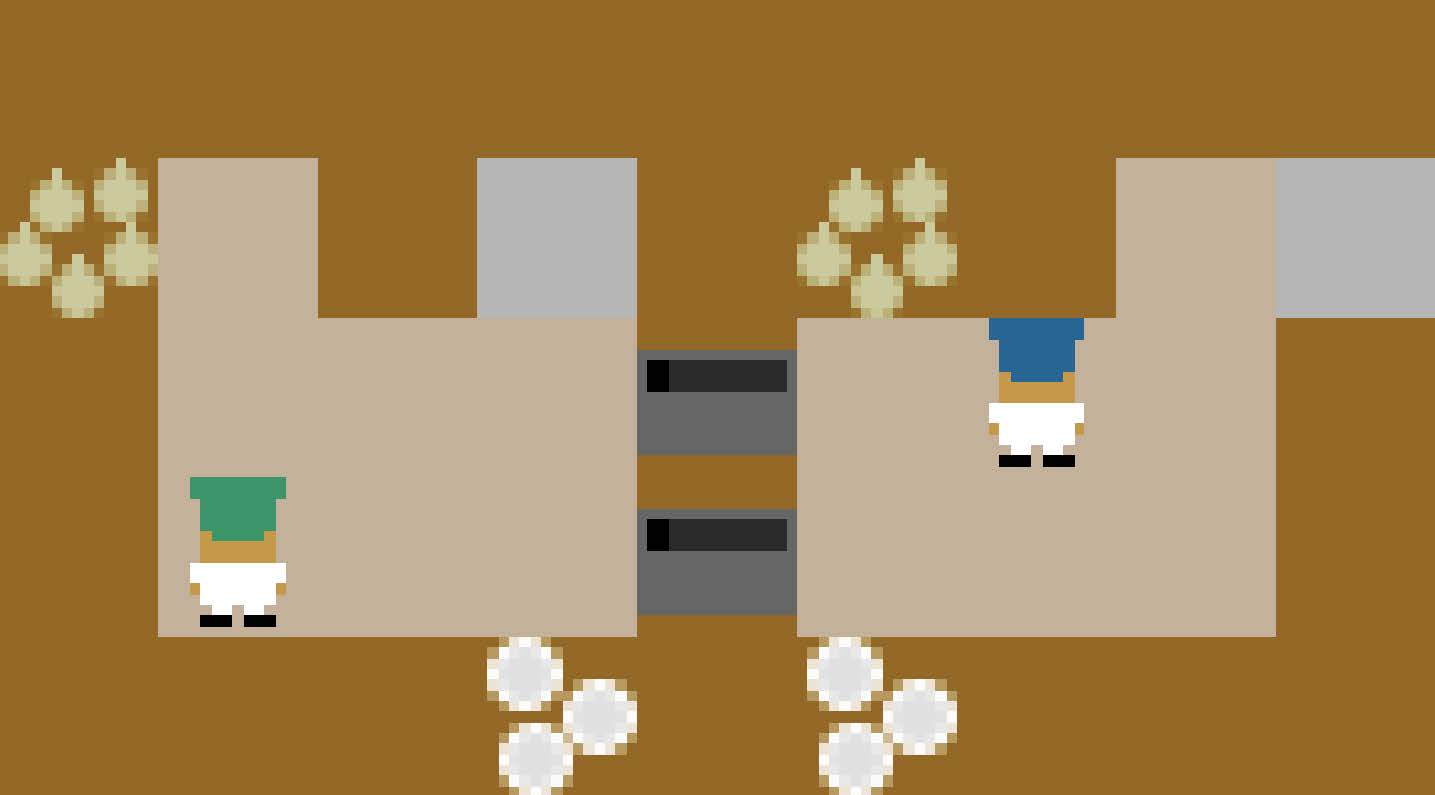}  
}   
\subfigure[{\scriptsize Coordination Ring}] {   
    \includegraphics[height=65pt]{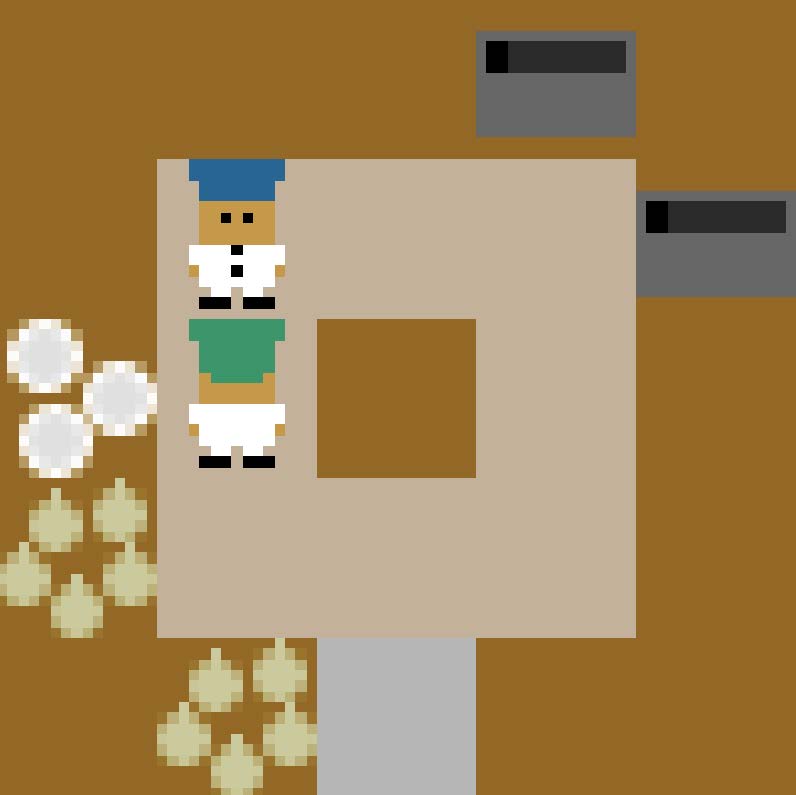}  
}   
\subfigure[{\scriptsize Forced Coordination}] {   
    \includegraphics[height=65pt]{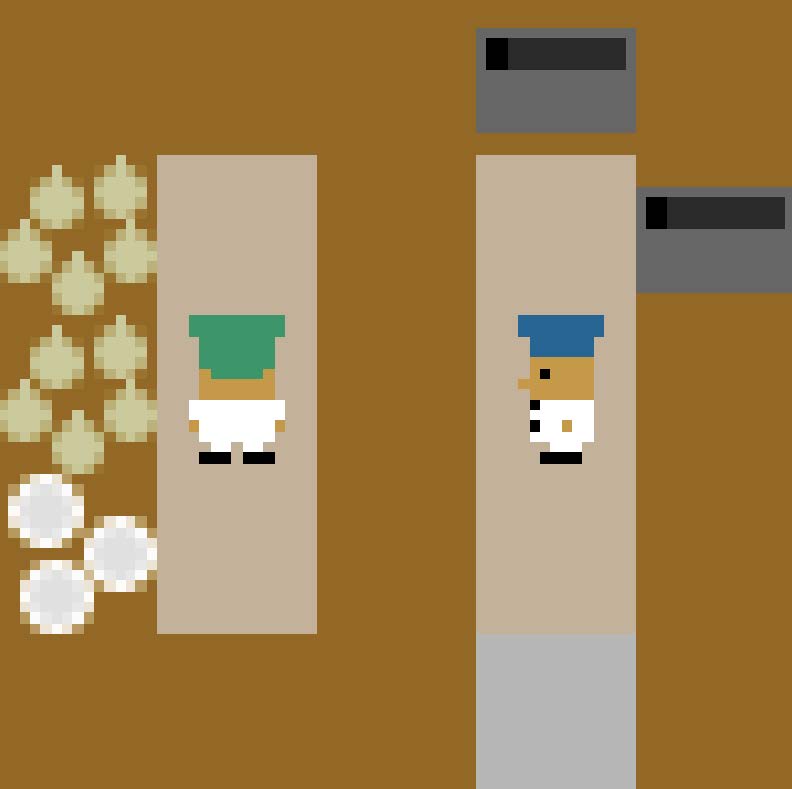}  
}   
\subfigure[{\scriptsize Counter Circuit}] {   
    \includegraphics[height=65pt]{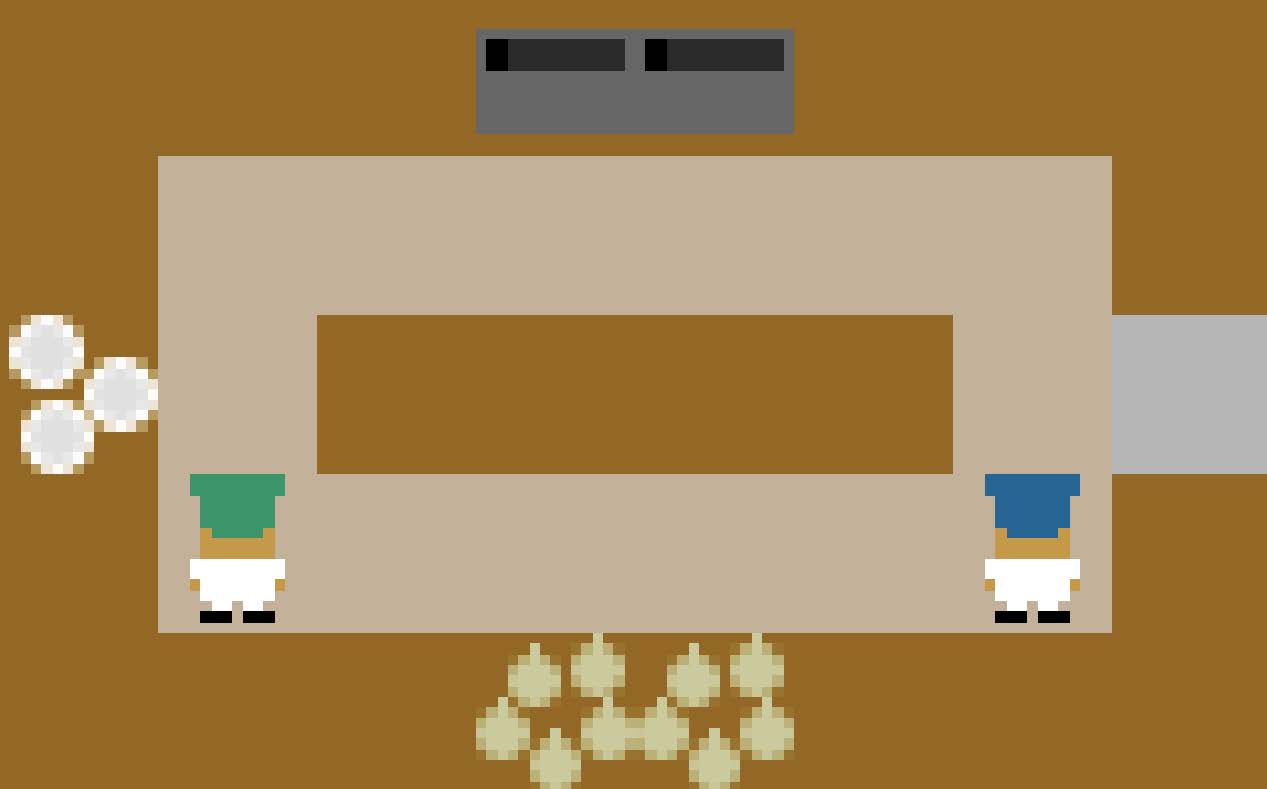}  
}   

\caption{ Overcooked environment layouts.}     
\label{fig:overcooked-layouts}     
\end{figure}

The Overcooked environment that we used has five layouts, including \textbf{Cramped Room}, \textbf{Asymmetric Advantages}, \textbf{Coordination Ring}, \textbf{Forced Coordination}, and \textbf{Counter Circuit}. Screenshots of these layouts can be seen in Fig.~\ref{fig:overcooked-layouts}.

The detailed introduction of five layouts is as follows.
\begin{enumerate}[label=(\alph*)]
\vspace{-3mm}
\setlength{\itemsep}{1pt}
\setlength{\parskip}{1pt}
\setlength{\parsep}{1pt}
    \item \textbf{Cramped Room}. The cramped room is a simple environment where two players are limited to a small room with only one pot (black box with gray bottom) and one serving spot (light gray square). Therefore, players are expected to fully utilize the pot and effectively deliver soup, even with basic coordination.

\item \textbf{Asymmetric Advantages}. In this layout, two players are placed in two disconnected kitchens. As the name suggests, the positions of onions, pots, and serving spots are asymmetric. In the left kitchen, onions are far from the pots, while serving spots are near the middle area of the layout. However, in the right kitchen, onions are placed near the middle area and the serving areas are far from the pots.

\item \textbf{Coordination Ring}. This ring-like layout requires both players to keep moving to prevent blocking each other, especially in the top-right and bottom-left corners where the onions and pots are located. For optimal cooperation, both pots should be utilized.

\item \textbf{Forced Coordination}. The Forced Coordination is another layout that separates the two agents. There are no pots or serving spots on the left side, nor are there onions or pots on the right side. Therefore, two players must coordinate with each other to complete the task. The left player is expected to prepare onions and plates while the right player cooks and serves them.

\item \textbf{Counter Circuit}. The Counter Circuit is another ring-like layout but larger in map size. In this layout, pots, onions, plates, and serving spots are placed in four different directions. Limited by the narrow aisles, players are easily blocked. Therefore, coordinating and performing the task is difficult in this environment. Players need to learn the advanced technique of putting onions in the middle area to pass them to the other quickly, which can further improve performance.        
\end{enumerate}

\section{Experimental Details of \algo}
\label{appendix:cole}
This paper utilizes Proximal Policy Optimization (PPO)~\citep{schulman2017proximal} as the oracle algorithm for our strategy set $\gN$, which consists of convolutional neural network parameterized strategies. Each network is composed of 3 convolution layers with 25 filters and 3 fully-connected layers with 64 hidden neurons. To manage computational resources, we maintain a population size of 50 strategies. In instances where the population exceeds this limit, we randomly select one of the earliest 10 strategies for removal.

We run and evaluate all our experiments on Linux servers, which include two types of nodes: 1) 1-GPU node with NVIDIA GeForce 3090Ti 24G as GPU and AMD EPYC 7H12 64-Core Processor as CPU, 2) 2-GPUs node with GeForce RTX 3090 24G as GPU and AMD Ryzen Threadripper 3970X 32-Core Processor as CPU.
On the Overcooked game environment, the \algo takes about one to two days on the 2-GPUs machine for one layout's training.

The hyperparameter setup is similar to those in PBT and MEP, which are given as follows. 
\begin{itemize}
    \item The learning rate for each layout is  2e-3 , 1e-3 , 6e-4 , 8e-4 , and 8e-4.
    \item The gamma $\gamma$ is 0.99.
    \item The lambda $\lambda$ is 0.98.
    \item The PPO clipping factor is 0.05.
    \item The VF coefficient is 0.5.
    \item The maximum gradient norm is 0.1.
    \item The total training time steps for each PPO update is 48000, divided into 10 mini-batches.
    \item The total numbers of generations for each layout are 80, 60, 75, 70, and 70, respectively.
    \item For each generation, we update 10 times to approximate the best-preferred strategy.
    \item The $\alpha$ is 1.
\end{itemize}

\section{Implementations of Baselines}
\label{appendix:base}
In this part, we will introduce the detailed implementations of baselines.
We train and evaluate the self-play and PBT based on the Human-Aware Reinforcement Learning repository~\citep{HARL} \footnote{\url{https://github.com/HumanCompatibleAI/human_aware_rl/tree/neurips2019}.} and used Proximal Policy Optimization (PPO)~\citep{schulman2017proximal} as the RL algorithm.
We implement FCP according to the FCP paper~\citep{FCP} and use PPO as the RL algorithm.
The implementation is based on the Human-Aware Reinforcement Learning repository (the same used in the self-paly and PBT).
The MEP agent is trained with population size as 5, following the MEP paper~\citep{MEP} and used the original implementation\footnote{The code of MEP original implementation: \url{https://github.com/ruizhaogit/maximum_entropy_population_based_training}.}.

\section{Trajectory Visualization}\label{Trajectory}
\begin{figure}[h]
    \centering
\includegraphics[width=0.75\linewidth]{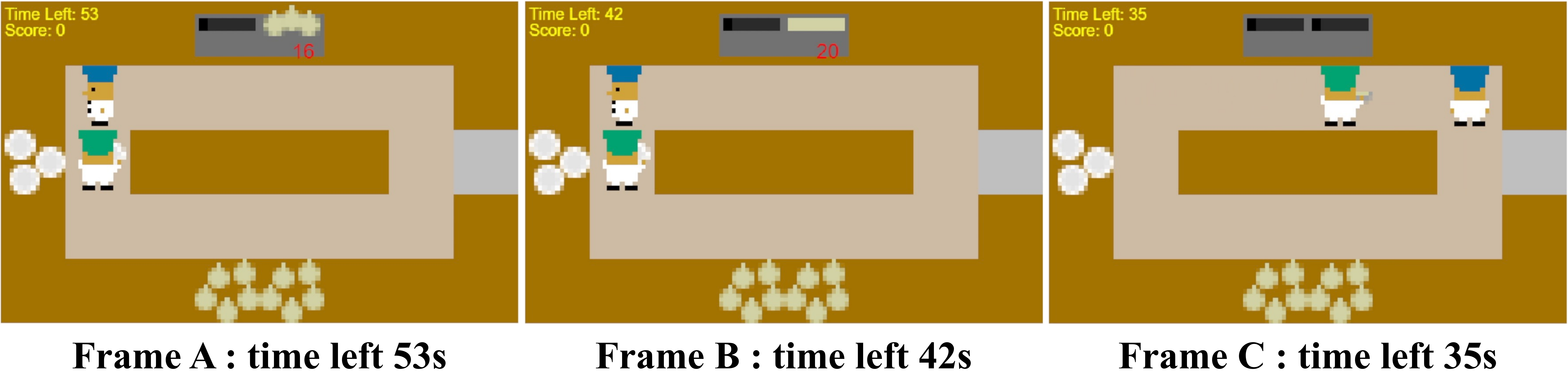}
\caption{
{Trajectory snapshots of the \algo 0:4 model (blue) with one of the expert partners - PBT model (green).}
}
    \label{fig:case_study}
\end{figure}

We visualize the trajectories produced by \algo 1:3 and 0:4 with middle-level and expert partners in Overcooked at \url{https://sites.google.com/view/cole-2023/}.
Fig.~\ref{fig:case_study} presents three screenshots of the \algo 0:4 model (blue player) that collaborates with one of the expert partners, the PBT model (green player). 
The case illustrates the importance of the individual objective in zero-shot coordination with expert partners. 
Frame A is a screenshot taken at 53s when the two players start to impede each other. 
The PBT model has taken the plate and wants to load and serve the dish. 
The blue player wants to take the plate but does not know how to change the objective to allow the green player to load the dish. 
After blocking for about 11s, the blue player starts to move and lets the green player go to the pots (Frame B). 
However, the process is not smooth and takes 7s to reach Frame C. 
This phenomenon does not occur in \algo 1:3 coordination with expert partners, which shows that including individual objectives might improve the cooperative ability with expert partners.

\end{document}